\documentclass[runningheads]{llncs}
\usepackage[T1]{fontenc}
\usepackage{amsmath}
\usepackage{booktabs}
\usepackage{numprint}
\usepackage{hyperref}

\usepackage{tikz}

\usepackage{graphicx}
\usepackage{color}

\begin{document}
\title{Fashion CUT: Unsupervised domain adaptation for visual pattern classification in clothes using synthetic data and pseudo-labels}
\titlerunning{Fashion CUT}
%
\author{Enric Moreu\inst{1,2}\orcidID{0000-0002-0555-3013} \and
Alex Martinelli\inst{1} \and
Martina Naughton\inst{1} \and
Philip Kelly\inst{1} \and
Noel E. O'Connor\inst{2}\orcidID{0000-0002-4033-9135}}
\authorrunning{E. Moreu et al.}
%
\institute{Zalando SE, Valeska-Gert-Straße 5, 10243 Berlin, Germany \and
Insight Centre for Data Analytics, Dublin City University, Dublin, Ireland}

\maketitle              
\begin{abstract}
Accurate product information is critical for e-commerce stores to allow customers to browse, filter, and search for products. Product data quality is affected by missing or incorrect information resulting in poor customer experience. While machine learning can be used to correct inaccurate or missing information, achieving high performance on fashion image classification tasks requires large amounts of annotated data, but it is expensive to generate due to labeling costs. One solution can be to generate synthetic data which requires no manual labeling. However, training a model with a dataset of solely synthetic images can lead to poor generalization when performing inference on real-world data because of the domain shift. We introduce a new unsupervised domain adaptation technique that converts images from the synthetic domain into the real-world domain. Our approach combines a generative neural network and a classifier that are jointly trained to produce realistic images while preserving the synthetic label information. We found that using real-world pseudo-labels during training helps the classifier to generalize in the real-world domain, reducing the synthetic bias. We successfully train a visual pattern classification model in the fashion domain without real-world annotations. Experiments show that our method outperforms other unsupervised domain adaptation algorithms. 



\end{abstract}
\section{Introduction}\label{section1}
\begin{figure}
\centering
\includegraphics[width=90px]{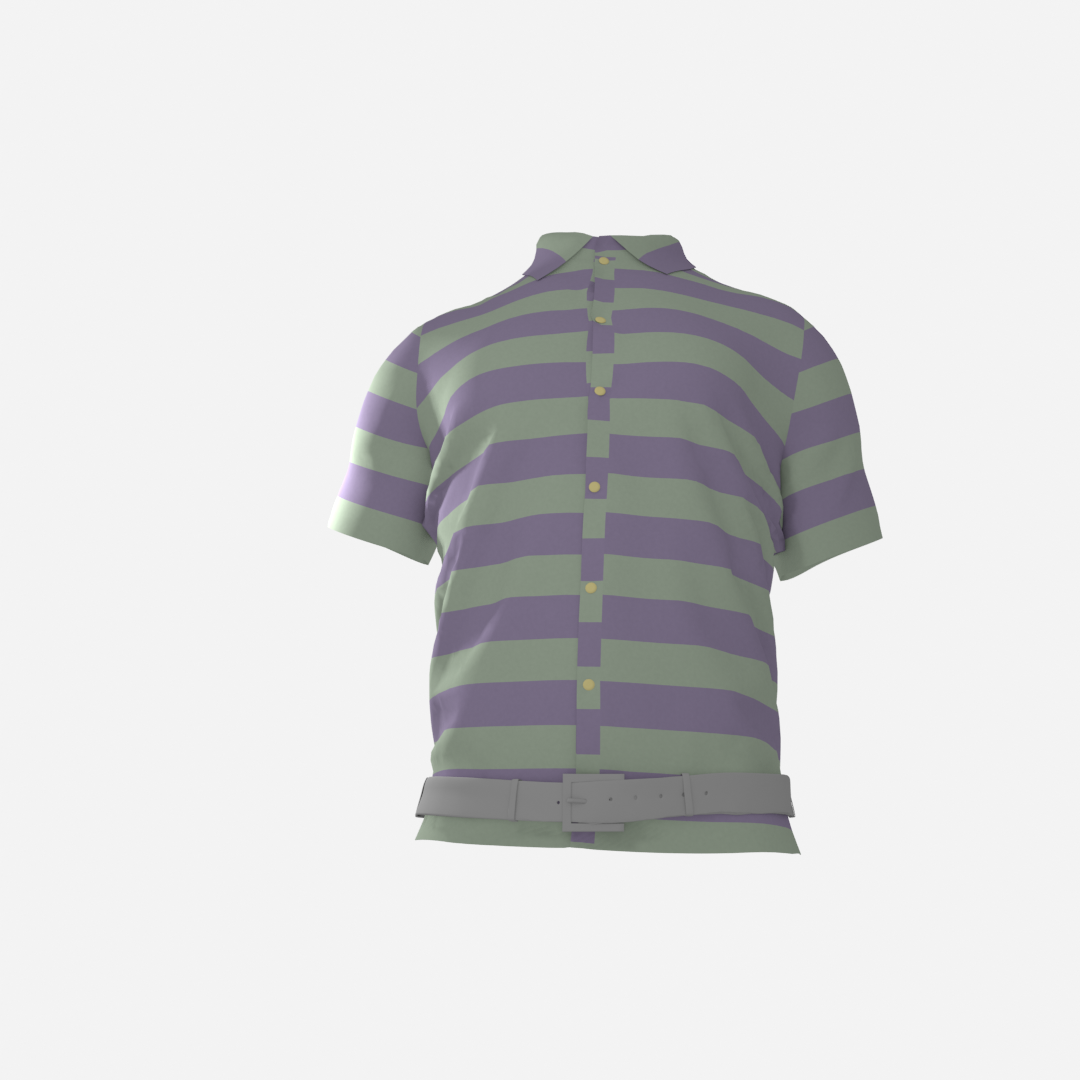}
\includegraphics[width=90px]{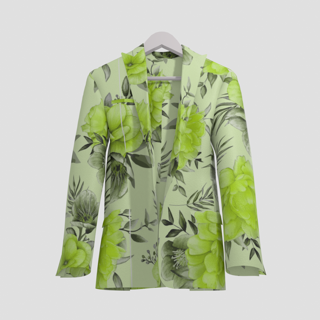}
\includegraphics[width=90px]{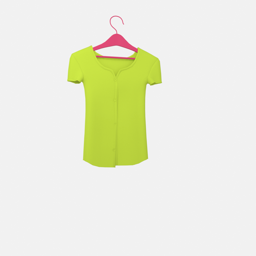}
\includegraphics[width=90px]{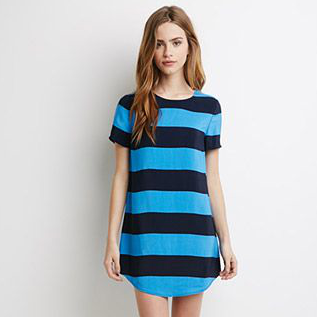}
\includegraphics[width=90px]{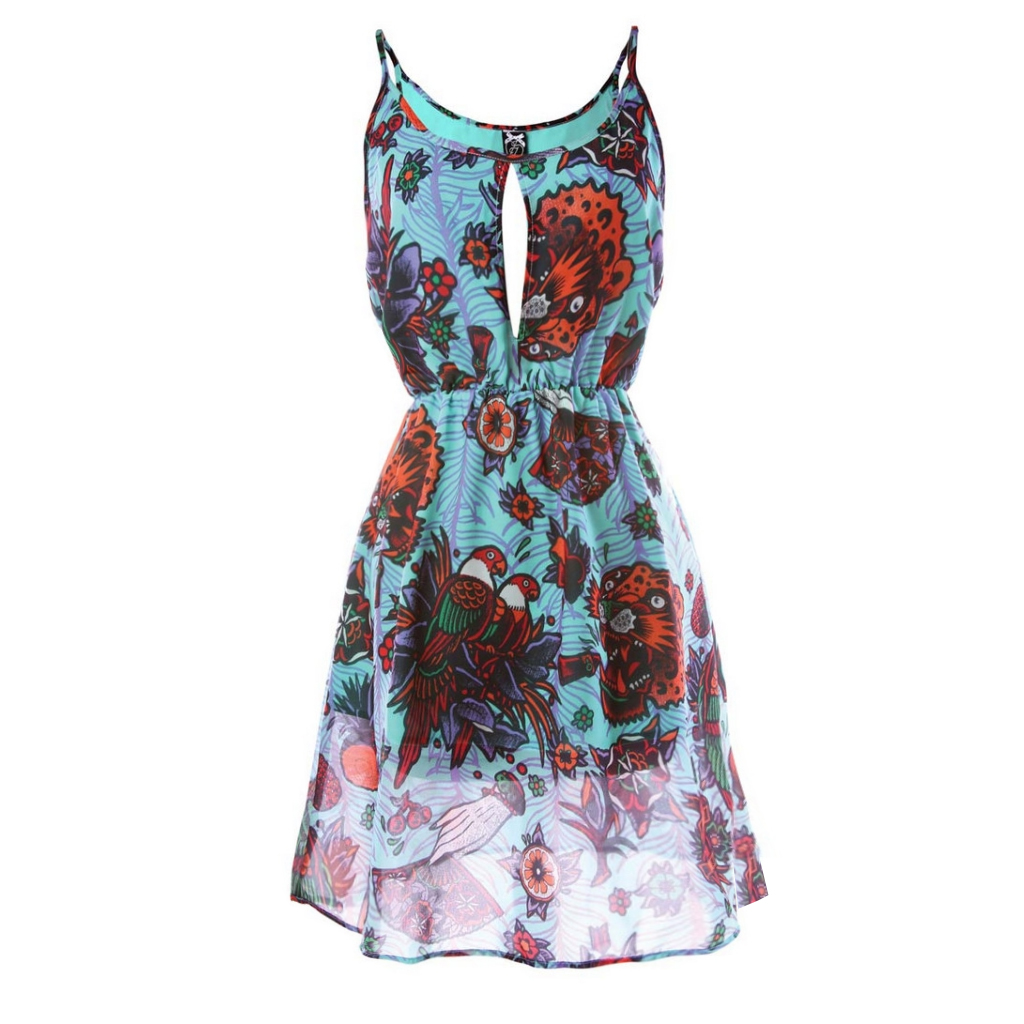}
\includegraphics[width=90px]{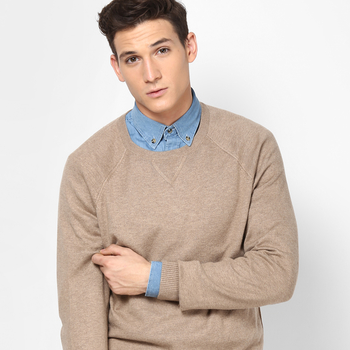}
\caption{Synthetic samples from our Zalando SDG dataset (first row) and real samples from the DeepFashion dataset~\cite{liuLQWTcvpr16DeepFashion} (second row) representing the striped, floral, and plain categories.}
\label{synth_vs_real}
\end{figure}

In 2021, 75\% of EU internet users bought goods or services online~\cite{lone20212021}. One of the main drivers of increased e-commerce engagement has been convenience, allowing customers to browse and purchase a wide variety of categories and brands in a single site.
If important product metadata is either missing or incorrect, it becomes difficult for customers to find products as the number of available products on e-commerce sites grows. 
Online stores typically offer a set of filters (e.g. pattern, color, size, or sleeve length) that make use of such metadata and help customers to find specific products. If such critical information is missing or incorrect then the product cannot be effectively merchandised. Machine learning has been used for fashion e-commerce in recent works to analyze product images, e.g. clothes retrieval~\cite{liang2016clothes}, detecting the outline~\cite{liu2016fashion}, or to find clothes that match an outfit~\cite{jagadeesh2014large}. In this paper, our prime interest is a  visual classification task which consists of classifying patterns in catalog images of clothing. Patterns describe the decorative design of clothes, and they are important because they are widely used by customers to find products online. Figure \ref{synth_vs_real} shows fashion visual pattern examples in the syntetic and real-world domains.

Fashion pattern classification is challenging. Fashion images often include models in different poses with complex backgrounds. Achieving high performance requires large annotated datasets~\cite{liuLQWTcvpr16DeepFashion}\cite{rostamzadeh2018fashion}\cite{guo2019fashion}. However, public datasets are only available for non-commercial use or do not cover the specific attributes or diversity we require, while generating private datasets with fine-grained and balanced annotations is expensive. In addition, publicly available fashion datasets typically have underrepresented classes with only a few samples. For example, in the Deep Fashion dataset~\cite{liuLQWTcvpr16DeepFashion} there are 6633 images with the “solid” pattern while only 242 images contain the “lattice” pattern. Categories that are underrepresented during training achieve a lower performance, thus reducing the overall performance.

We address these problems by generating artificial samples using Synthetic Data Generation (SDG) techniques. Synthetic data has shown promising results in domains where few images are available for training~\cite{sankaranarayanan2018learning}\cite{moreu2022joint}. 
The main advantage of synthetic data is that it can generate unlimited artificial images because labels are automatically produced by the 3D engine when rendering the images.

However, synthetic images are not a precise reflection of the real-world domain in which the model will operate. Computer vision models are easily biased by the underlying distribution in which they are trained~\cite{nam2021reducing}. Even if the synthetic images use realistic lighting and textures that look realistic to humans, the model will tend to over-optimize against the traits of the synthetic domain and won’t generalize well to real data.

We consider this problem in the context of unsupervised domain adaptation~\cite{wang2018deep} by using the knowledge from another related domain where annotations are available. We assume that we have abundant annotated data on the source domain (synthetic images) and a target domain (real images) where no labels are available. 

Unsupervised domain adaptation has shown excellent results when translating images to other domains~\cite{wang2018deep} ; nevertheless, translated images can’t be readily used to train classification models because image features, such as patterns, are distorted during the translation step since the translation model doesn’t have information about the features.
 Specifically, when complex patterns are shifted to a different domain, they can be distorted to a level that they no longer adhere to the original pattern label for the synthetic image. For example, when an image with the “camouflage” pattern is translated from the synthetic to the real domain, the pattern could be accidentally distorted to “floral”.
 
In this paper, we introduce a new unsupervised domain adaptation approach that doesn’t require groundtruth labels. First, we produce a synthetic dataset for fashion pattern classification using SDG that equally represents all the classes. Second, we jointly train a generative model and a classifier that will make synthetic images look realistic while preserving the class patterns. In the final stage of the training, real-world pseudo-labeled images are used to improve the model generalization toward real images. The contributions of this paper are as follows:
\begin{itemize}
    \item We propose a novel architecture that performs the image translation task while jointly training a classification model.
    \item We outperform other state-of-the-art unsupervised domain adaptation algorithms in the visual fashion pattern classification task.
\end{itemize}

The remainder of the paper is organized as follows: Section 2 reviews relevant work; Section 3 explains our method;  Section 4 presents our synthetic dataset and experiments, and Section 5 concludes the paper. 

\section{Related work}\label{section2}
Synthetic data has been used extensively in the computer vision field. Techniques to generate synthetic datasets range from simple methods generating primitive shapes~\cite{rahnemoonfar2017deep} to photorealistic rendering using game engines~\cite{wang2019learning}. 
Although high quality synthetic images can appear realistic to humans, they don't necessarily help the computer vision models to generalize to real-world images. Convolutional neural networks easily overfit on synthetic traits that are not present in the real-world.
This is addressed by using domain adaptation techniques that reduces the disparity between the synthetic and real domains. Some works approach domain adaptation by simply improving the realism aspect~\cite{ros2016synthia}, or by pushing the randomization and distribution coverage at the source~\cite{moreu2022domain}. These approaches imply additional modeling effort and longer generation times per image, for example by relying on physically based renderers for higher photorealistic results, making the synthetic data better match the real data distribution.

\paragraph{}
In the context of unsupervised domain adaptation, non-adversarial approaches consist of matching feature distributions in the source and the target domain by transforming the feature space to map the target distribution~\cite{Xu_2019_ICCV}. Gong et al.~\cite{gong2012geodesic} found that gradually shifting the domains during training improved the method’s stability. Recent methods are based on generative adversarial networks~\cite{goodfellow2020generative} because of their unsupervised and unpaired nature. Generative domain adaptation approaches rely on a domain discriminator that distinguishes the source and target domains~\cite{ganin2016domain} and updates the generator to produce better images. Our approach improves existing adversarial approaches by optimizing a classifier alongside the generator, producing realistic data that retrain the source category.

\section{Fashion CUT}\label{section3}
\begin{figure}
\centering
\includegraphics[width=350px]{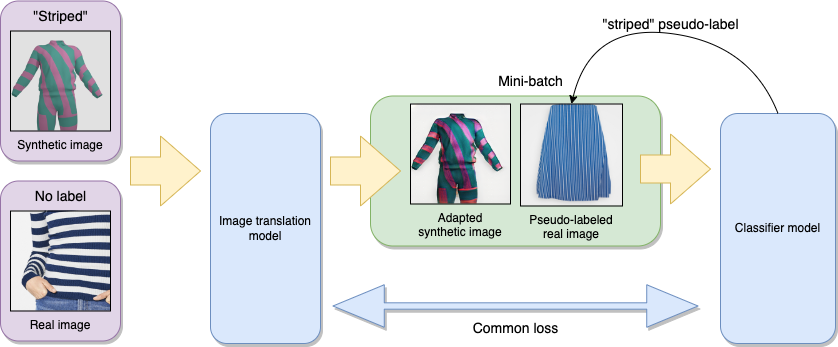}
\caption{The proposed architecture includes a translation model (CUT) and a classifier model (ResNet50), which are optimized together via a common loss that ensures realistic images with reliable annotations. Pseudo-labeled real images are included in each mini-batch to improve the classifier generalization.}
\label{arch}
\end{figure}

Our approach has two components: 1) An image translation network that generates realistic images. 2) A classifier that enforces the generated images to keep the class patterns. The overall architecture is shown in Figure \ref{arch}.

Acquiring paired images from both domains can be difficult to achieve in the fashion domain, resulting in high costs. As such, we use Contrastive Unpaired Translation (CUT)~\cite{park2020cut} for the image translation module. Synthetic images don’t have to match the exact position or texture of real images in the dataset because we use an unpaired translation method. CUT learns a mapping that translates unpaired images from the source domain to the target domain. It operates on patches that are mapped to a similar point in learned feature space using an infoNCE~\cite{gutmann2010noise} contrastive loss.
In addition, CUT uses less GPU memory than other two-sided image translation models (e.g. CycleGAN) because it only requires one generator and one discriminator. By reducing memory usage, the joint training of an additional classifier becomes tractable on low cost GPU setups with less than 16GB of memory.

\begin{figure}
\centering
\includegraphics[width=90px]{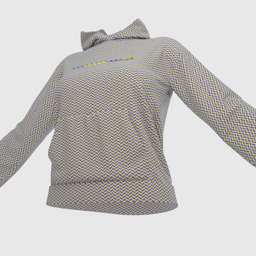}
\includegraphics[width=90px]{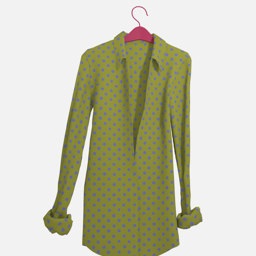}
\includegraphics[width=90px]{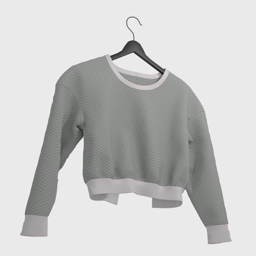}
\includegraphics[width=90px]{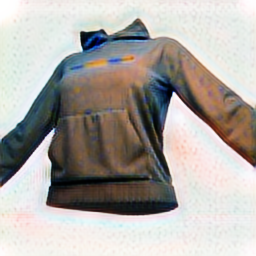}
\includegraphics[width=90px]{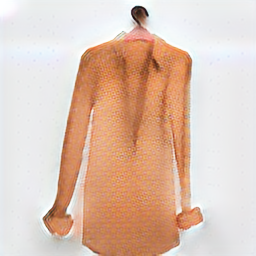}
\includegraphics[width=90px]{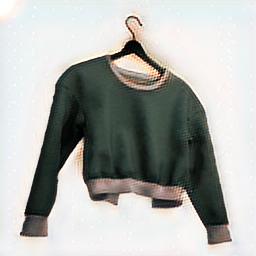}
\caption{ Synthetic images (first row) and unsuccessfully adapted images using CUT (second row) due to shifted patterns by the generator when not imposing class constraints.}
\label{cut_translation}
\end{figure}

\begin{figure}
\centering
\includegraphics[width=90px]{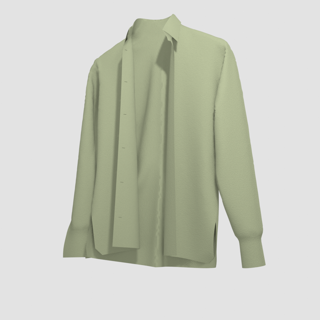}
\includegraphics[width=90px]{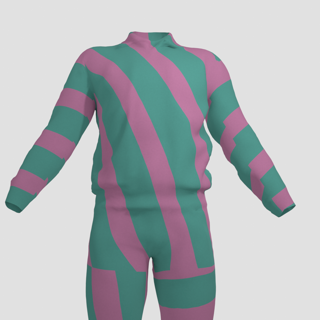}
\includegraphics[width=90px]{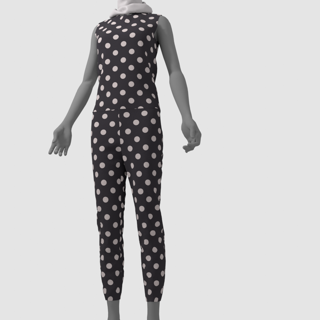}
\includegraphics[width=90px]{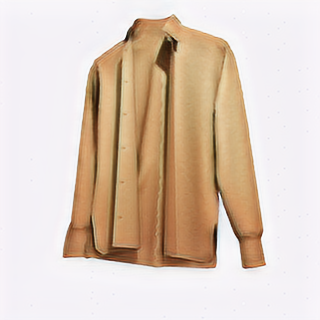}
\includegraphics[width=90px]{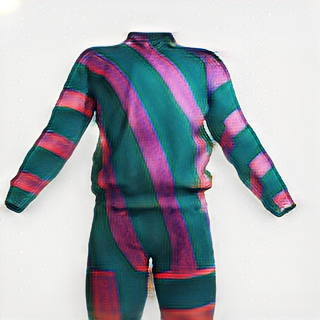}
\includegraphics[width=90px]{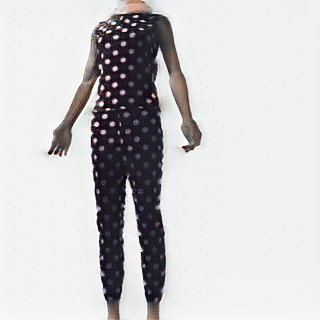}
\caption{ Synthetic images (first row) and adapted domain images using Fashion CUT(second row).}
\label{fcut_translation}
\end{figure}

While CUT produces realistic images, the class patterns can be lost or mixed with other classes since CUT doesn’t enforce that these category features are consistent across the image translation. The generator's only objective is to produce realistic images that resemble the real-world domain, but it ignores the nature of each pattern. Any pattern distorted during the translation will impact the performance of a classifier trained on this synthetic data. Figure \ref{cut_translation} showcases unsuccessful examples of mixed patterns by the generator, and Figure \ref{fcut_translation} shows successful translations using Fashion CUT.

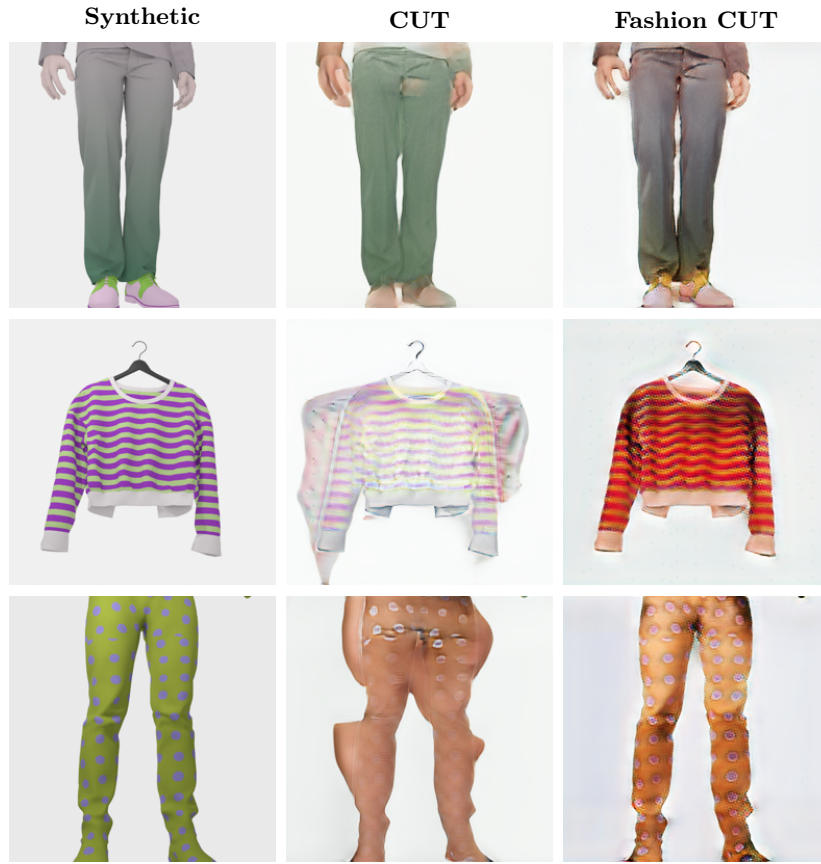
\begin{figure}[t!]
\centering
\begin{tikzpicture}[picture format/.style={inner sep=2pt,}]

  \node[picture format]                   (A1)               {\includegraphics[width=100px]{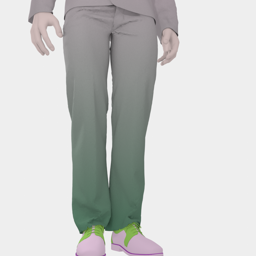}};
  \node[picture format,anchor=north]      (B1) at (A1.south) {\includegraphics[width=100px]{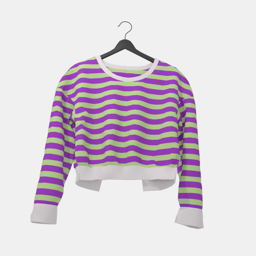}};
  \node[picture format,anchor=north]      (C1) at (B1.south) {\includegraphics[width=100px]{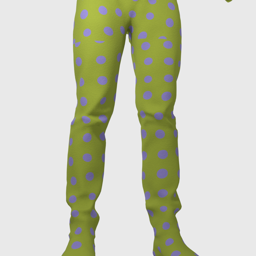}};

  \node[picture format,anchor=north west] (A2) at (A1.north east) {\includegraphics[width=100px]{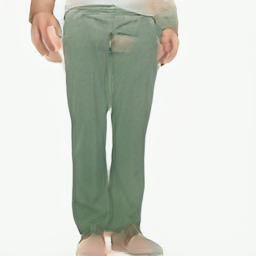}};
  \node[picture format,anchor=north]      (B2) at (A2.south)      {\includegraphics[width=100px]{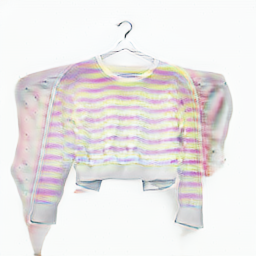}};
  \node[picture format,anchor=north]      (C2) at (B2.south)      {\includegraphics[width=100px]{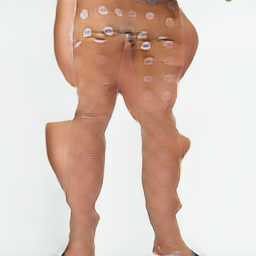}};

  \node[picture format,anchor=north west] (A3) at (A2.north east) {\includegraphics[width=100px]{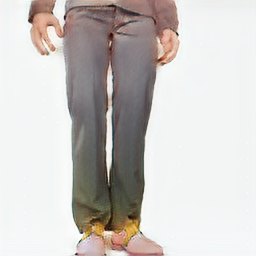}};
  \node[picture format,anchor=north]      (B3) at (A3.south)      {\includegraphics[width=100px]{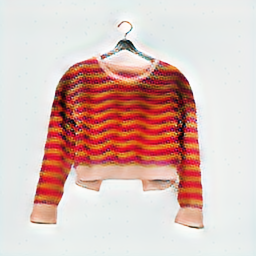}};
  \node[picture format,anchor=north]      (C3) at (B3.south)      {\includegraphics[width=100px]{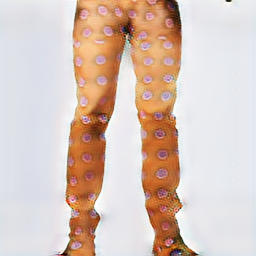}};


  \node[anchor=south] (D1) at (A1.north) {\bfseries Synthetic};
  \node[anchor=south] (D2) at (A2.north) {\bfseries CUT};
  \node[anchor=south] (D3) at (A3.north) {\bfseries Fashion CUT};

\end{tikzpicture}
\caption{Comparison of CUT and Fashion CUT image translation. Note that the annotations (gradient, striped, dotted) are preserved when using Fashion CUT.}
\label{cut_vs_fcut}
\end{figure}

In order to enforce stability in the generated patterns, we add a ResNet50 model that predicts the category of the images generated by CUT. The classifier is optimized alongside the CUT generator to fulfill both classification and translation tasks. Figure \ref{cut_vs_fcut} shows how the classifier preserves the pattern features in comparison to vanilla CUT. Training both models simultaneously is faster and provides better results than training them separately.
The generator loss function is given by:

\begin{equation}
\begin{split} \lambda{g}*\mathcal{L}_\mathit{GAN}(\mathit{G, D, X, Y }) + \lambda{c}*\mathcal{L}_\mathit{classifier}(\mathit{C}) + \\
     \lambda{ncex}*\mathcal{L}_\mathit{NCEx}(\mathit{G, D, X }) + \lambda{ncey}*\mathcal{L}_\mathit{NCEy}(\mathit{G, D, Y })
\end{split}
\end{equation}
\label{loss}

where $\mathcal{L}_\mathit{GAN}(\mathit{G, D, X, Y })$ is the generator loss, $\mathcal{L}_\mathit{classifier}(\mathit{C})$ is the cross-entropy loss on the classifier inferred from the images generated by the generator. $\mathcal{L}_\mathit{NCEx}(\mathit{G, D, X })$ and $\mathcal{L}_\mathit{NCEy}(\mathit{G, D, Y })$ are the contrastive losses that encourage spatial consistency for the synthetic and real images, respectively. $\mathit{G}$ is the generator model, $\mathit{D}$ the discriminator model, $\mathit{X}$ the real image, $\mathit{Y}$ the synthetic image, and $\mathit{C}$ the classification model. $\lambda{g}$, $\lambda{c}$, $\lambda{ncex}$, and $\lambda{ncey}$ are hyperparameters that control the weight of the generator, the classifier, and both contrastive losses, respectively.

In our experiments we empirically choose to replace half of the synthetic mini-batch with images from the target domain. As real-world annotations are not available for generated images, we use pseudo-labels predicted by the classifier. The model suffers from the cold start problem when introducing pseudo-labels in the early epochs because the classifier struggles to converge. We found that the classifier requires at least 1 epoch of synthetic samples in order to generate reliable pseudo-labels for real-world images. We obtained the best results when enabling pseudo-labels at the end of epoch 2. 

\section{Experiments}\label{section4}
This section describes the synthetic dataset we generated and the two experiment setups used to evaluate Fashion CUT.
\subsection{Zalando SDG dataset}
\begin{figure}
\centering
\includegraphics[width=250px]{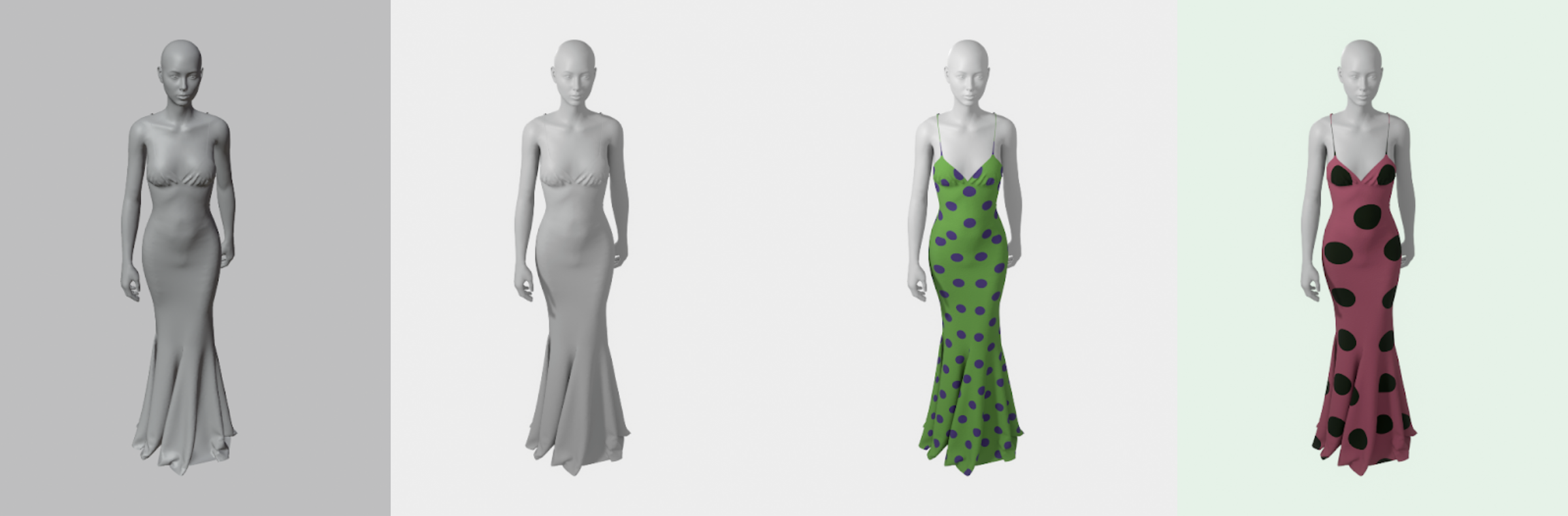}
\caption{For each render we start with a provided 3D object, add environment and spot lights, apply a procedural material and then randomize its properties (e.g. colors, scale).}
\label{blender_image}
\end{figure}

The Zalando SDG dataset is composed of 31,840 images of 7 classes: plain, floral, striped, dotted, camouflage, gradient, and herringbone. The dataset has been generated using Blender, an open-source 3D computer-graphic software~\cite{blender}. We relied on a basic set of professionally modeled 3D objects from CGTrader representing a variety of fashion silhouettes (e.g. shirt, dress, trousers) and implemented a procedural material for each of the 7 target classes. Each procedural material is implemented as a Blender shader node, where multiple properties can be exposed and controlled via Blender Python API. Examples of such properties include pattern scale, color or color pairing, orientation and image-texture. This setup allows an arbitrary amount of different images for each 3D object and class pair to be generated programmatically. We randomized background, lighting, and camera position, as seen in Figure \ref{blender_image}. We didn't use physically based renderers as those are more resource intensive, instead we traded off rendering accuracy for speed and adopted the real-time Blender Eevee render engine~\cite{guevarra2019modeling}.
 
The procedural materials can be applied to any new 3D objects. As such they provide a powerful generalized approach to data creation, and the generated images do not require any manual human validation as long as the procedural randomization guarantees that each possible output belongs to the expected target domain class.

\subsection{Evaluation on Zalando SDG dataset}

In our experiments we train end-user pattern classification models using datasets from both 
31,840 synthetic fashion imagery (the source domain, which includes groundtruth labels), and 334,165 real-world fashion imagery (the target domain, which has no groundtruth labels and is used solely to train our domain adaptation transformation).
We evaluate the performance of the algorithms using a validation set and a test set composed of 41,667 annotated real images each. The metric used is accuracy and all algorithms use a ResNet50~\cite{he2016deep} as the classifier. Fashion CUT is optimized using Adam with learning rate $10^{-5}$ and $\lambda{g}=0.1$, and $\lambda{classifier}=0.1$ for $\mathit{N}=5$ epochs. 

In Table \ref{results}, we compare the performance of domain adaptation algorithms trained only on our 31,840 synthetically generated dataset and evaluated on the 41,667 real fashion images. 

\begin{table}[t]
\centering
\begin{tabular}{lc}
\toprule
Method &  Accuracy\\
\midrule
No adaptation & 0.441\\
BSP~\cite{chen2019transferability}  & 0.499\\
MDD~\cite{zhang2019bridging} & 0.540\\
AFN~\cite{Xu_2019_ICCV} & 0.578\\
Fashion CUT (ours) & 0.613 \\
Fashion CUT with pseudo-labels (ours) & \textbf{0.628}\\
\bottomrule
\end{tabular}
\caption{Comparison of unsupervised domain adaptation algorithms on our Zalando SDG dataset. The metric used is accuracy.}\label{results}
\end{table}

First, we measured the performance of training without domain adaptation. In other words, the classifier was trained only on synthetic images. The performance was poor because the model didn’t have information about real world images.

Second, we evaluate Zalando SDG on other domain adaptation algorithms in the fashion domain. All experiments were performed in the environment provided by Jiang et al.~\cite{tllib}. Our approach outperforms the other algorithms for the pattern classification task. Finally, we found that using pseudo-labels improves the results with minor changes in the training.

\subsection{Synthetic dataset size}
\begin{figure}
\centering
\includegraphics[width=200px]{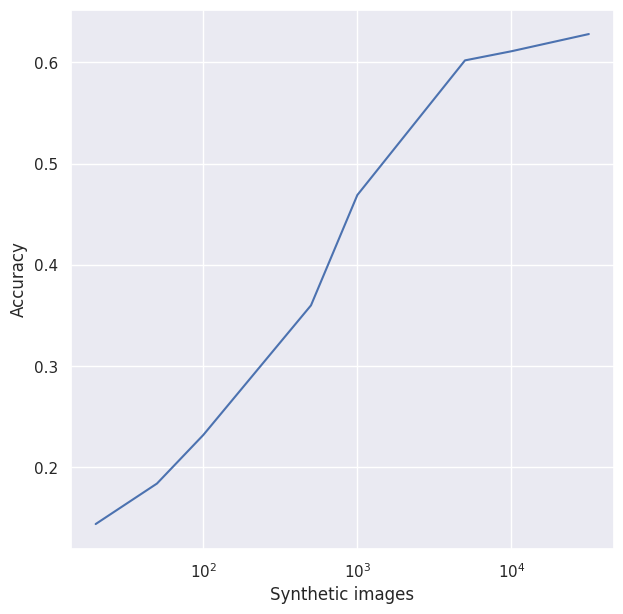}
\caption{Evaluation of Fashion Cut with varying amounts of the Zalando SDG dataset and 10.000 unlabeled real images. The performance is measured in accuracy.}
\label{synth_size}
\end{figure}

We explore the required number of synthetic images to successfully train our unsupervised domain adaptation algorithm. For this experiment we train our model using 10,000 unlabeled real images and changing the number of synthetic images. Figure \ref{synth_size} shows that Fashion CUT performance benefits from large synthetic datasets. We found that at least 5,000 synthetic images are required to outperform other algorithms in visual pattern classification.

\section{Conclusions}\label{section5}
Combining synthetic data generation with unsupervised domain adaptation can successfully classify patterns in clothes without real-world annotations. Furthermore, we found that attaching a classifier to an image translation model can enforce label stability, thus improving performance. Our experiments confirm that Fashion CUT outperforms other domain adaptation algorithms in the fashion domain. In addition, pseudo-labels proved to be beneficial for domain adaptation in the advanced stages of the training. As future work, we will explore the impact of fashion synthetic data in a semi-supervised setup. We hope this study will help enforce 3D rendering as a replacement for human annotations. 

\bibliographystyle{splncs04}
\bibliography{citations}

\end{document}